\documentclass[double]{article}
\usepackage[T1]{fontenc}
\usepackage{times}
\usepackage{algorithm}
\usepackage{algorithmic}
\usepackage{wrapfig}
\usepackage{verbatim}
\usepackage{amsmath}
\usepackage{graphicx}
\usepackage{subcaption}
\usepackage[active]{srcltx}
\usepackage{color}
\usepackage{lineno}
\usepackage{dirtytalk}
\usepackage{amsthm}
\usepackage{authblk}
\usepackage{listings}
\usepackage{tikz}
\usepackage{longtable}
\usepackage{setspace}
\usepackage[backend=biber,style=numeric,sorting=none,maxnames=9,minnames=1]{biblatex}
\usetikzlibrary{shapes.geometric, arrows}

\usepackage{epsfig,graphicx,amsfonts}
\usepackage{amsmath,amsthm,amssymb}
\usepackage{xcolor}

\usepackage{adjustbox}
\usepackage{multirow}
\usepackage{setspace}
\usepackage{hyperref}
\usepackage{comment}

\long\def\comment #1\commentend{}

\begin{document}

\title{\Large Accelerating Reinforcement Learning Algorithms Convergence using Pre-trained Large Language Models as Tutors With Advice Reusing}

\author{Lukas Toral$^{1}$, Teddy Lazebnik$^{2,1,*}$\\ \(^1\) Department of Computing, Jonkoping University, Jonkoping, Sweden \\ \(^2\) Department of Information Science,  University of Haifa, Haifa, Israel \\ \(^*\) Corresponding author: lazebnik.teddy@gmail.com \\ }

\date{ }

\maketitle 

\begin{abstract}
Reinforcement Learning (RL) algorithms often require long training to become useful, especially in complex environments with sparse rewards. While techniques like reward shaping and curriculum learning exist to accelerate training, these are often extremely specific and require the developer's professionalism and dedicated expertise in the problem's domain. Tackling this challenge, in this study, we explore the effectiveness of pre-trained Large Language Models (LLMs) as tutors in a student-teacher architecture with RL algorithms, hypothesizing that LLM-generated guidance allows for faster convergence. In particular, we explore the effectiveness of reusing the LLM's advice on the RL's convergence dynamics. Through an extensive empirical examination, which included 54  configurations, varying the RL algorithm (DQN, PPO, A2C), LLM tutor (Llama, Vicuna, DeepSeek), and environment (Blackjack, Snake, Connect Four), our results demonstrate that LLM tutoring significantly accelerates RL convergence while maintaining comparable optimal performance. Furthermore, the advice reuse mechanism shows a further improvement in training duration but also results in less stable convergence dynamics. Our findings suggest that LLM tutoring generally improves convergence, and its effectiveness is sensitive to the specific task, RL algorithm, and LLM model combination. \noindent \\\\

\end{abstract}

\maketitle \thispagestyle{empty}
\pagestyle{myheadings} \markboth{Draft:  \today}{Draft:  \today}
\setcounter{page}{1}

\onehalfspacing

\section{Introduction}\label{sec:introduction}
Reinforcement Learning (RL) is a branch of machine learning where an agent learns to make decisions by interacting with an environment, mimicking the learning process of many biological organisms \cite{intro_1,intro_2,intro_3}. The agent receives feedback in the form of rewards or penalties based on its actions, and it aims to maximize cumulative rewards over time \cite{intro_4}. For instance, an autonomous robot navigating a maze receives a reward for reaching the exit efficiently and a penalty for colliding with walls, which it uses to learn an optimal strategy (or \say{policy}) \cite{intro_5}. In recent years, RL has emerged as a powerful paradigm in artificial intelligence (AI) for solving sequential decision-making problems \cite{intro_8,intro_9,intro_10}. From economics optimization \cite{intro_11}, game theory-inspired decision-making \cite{intro_12}, and autonomous systems like cars \cite{intro_13} and factories \cite{intro_14}, RL has demonstrated remarkable success in environments where explicit programming is challenging \cite{intro_15}. 

Despite these advances, RL algorithms often suffer from slow convergence, requiring extensive computational resources to reach (near-)optimal performance \cite{intro_16,intro_17,intro_18}. To emphasize this point, Deep Q-Networks (DQN), a specific instance of a RL model that is trained to play Atari games \cite{mnih2015human}, required around 200 million environment steps for training, which translates to several weeks on modern GPUs (Graphics processing units) compute time, which consider unreachable for many practitioners. This challenge becomes particularly pronounced in complex environments with sparse rewards, where agents struggle to learn effective policies efficiently \cite{intro_19}. 

To address this challenge, scholars and practitioners have explored various techniques to accelerate RL training \cite{intro_20,intro_21}. These approaches include reward shaping \cite{intro_22}, curriculum learning \cite{intro_23}, imitation learning \cite{intro_24}, and meta-learning \cite{intro_25}, each providing a different mechanism for improving sample efficiency and learning speed. More recently, large language models (LLMs) have demonstrated extraordinary capabilities in reasoning, generalization, and knowledge distillation, raise the possibility of leveraging them as external tutors to guide RL agents \cite{intro_28}. These models, such as GPT \cite{intro_27} or DeepSeek \cite{deepseekai2025deepseekr1incentivizingreasoningcapability}, can generate human-like text and be fine-tuned for specific tasks, and by being pre-trained on large volumes of data, can use external knowledge for a wide range of decision making tasks. Using pre-trained LLMs to provide structured guidance, explanations, and action recommendations introduces a novel way to facilitate RL convergence without requiring direct modifications to the underlying RL algorithms. 

Despite growing interest in integrating LLMs with RL, existing research primarily focuses on natural language-conditioned RL or learning from textual instructions \cite{intro_29}. These studies often fail to systematically investigate the role of LLMs as interactive tutors that actively guide an RL agent's learning process. In addition, current studies have not rigorously quantified how much LLM-assisted guidance impacts learning speed, especially under the repetitive consulting mechanism \cite{intro_30}. To this end, in this study, we explored the effectiveness of LLMs as tutors in a student-teacher architecture \cite{intro_31} where RL algorithms operate as students. We hypothesize that incorporating LLM-generated guidance allows RL agents to achieve faster convergence, particularly in sparse reward environments. To be exact, explored two main research questions: i) Do LLM tutors contribute to faster convergence of RL models to a better optimum?; and ii) does reusing advice given by the LLM tutors contribute to even faster convergence of RL models than just using LLMs? Through extensive empirical evaluations across 54 configurations of tasks, RL algorithms, and LLM models, we demonstrate that indeed LLMs accelerate RL convergence while not crucially altering the overall RL's optimal performance and that the reusing mechanism emphasizes this outcome while also introducing instability to the training process as a trade-off.

The rest of this manuscript is organized as follows. Section \ref{sec:rw} presents the current state-of-the-art in RL and LLMs as well as previous attempts to integrate the two. Next, section \ref{sec:methods} formally introduces the proposed LLM and RL integration mechanism, followed by the experimental setup. Afterward, section \ref{sec:results} outlines the obtained results. Finally, Section \ref{sec:discussion} discuss the results in an applied context and suggests possible future work. 

\section{Related Work}
\label{sec:rw}
In this section, we outline the computational methods used by our method, including the RL and LLM. Next, we review the current status of integrating LLM with RL and the reusing advice mechanism in this context. 

\subsection{Reinforcement Learning} \label{rlBackground}
RL is a powerful family of AI methods that can solve many complex and usually poorly-formally-defined optimization tasks \cite{sedationPaper,liu2022finrldeepreinforcementlearning,schultz2018deepreinforcementlearningdynamic} such as in fields like healthcare \cite{Poolla2003ARL,anesthesiaPaper}, finance \cite{evgeny_ponomarev__2019,8701368} transportation \cite{DBLP:journals/corr/abs-1802-04240,8967630}, to name a few. Formally, RL refers to a family of models that use learning functions (\(\pi: S \rightarrow A\)) that map between states (\(S\)) to actions (\(A\)) to maximize cumulative numerical reward over time \cite[p.~7-9]{suttonRLBook}. Equally important for the RL definition is the way such a mapping function is learned, which is inspired by biological organisms such as animals and humans \cite{rl_bio_1,rl_bio_2} in a process of repeated trial-and-error \cite[p.~9]{suttonRLBook}. This approach is considered very useful, as it is mathematically guaranteed to eventually converge to an optimal strategy, given enough time \cite{littman1996generalized}. 

There are other key components to a RL-based model, including a policy function (\(\pi\)), reward signal (\(R\)), value function (\(V\)), and optionally a model of the environment (\(M\)) \cite[p.~7]{suttonRLBook}. An agent's policy defines the agent's behavior based on the given state of the environment. For simple use-cases, a policy might be a lookup table or a linear function \cite[p.~7]{suttonRLBook}. Nevertheless, not all environments can be solved with such a policy due to their complexity (e.g., continuous observation space), and more complex techniques have to be used, such as approximation using a neural network \cite[p.~7]{suttonRLBook}. A reward signal represents the goal in an RL problem - after each agent's step in the environment, it receives a single number, called reward \cite[p.~7]{suttonRLBook}. Rewards are used to guide the agent's training process and are correlated to the desired goal that the RL-based model should achieve \cite[p.~7]{suttonRLBook}. The reward signal is the primary basis of changing the agent's policy - if the action selected by a policy yielded a low reward, the policy might have to be changed to select a different action in the future \cite[p.~7]{suttonRLBook}. The agent uses the value function to predict the reward it would receive by executing a certain action \cite[p.~7]{suttonRLBook}. It is, however, more challenging to determine values than it is to determine a reward, as rewards are given by an environment directly, and values must be estimated by executing actions \cite[p.~8]{suttonRLBook}. Then, finally, the model of the environment allows inferences about how the environment will behave \cite[p.~8]{suttonRLBook}. To this end, the RL environments can more often than not be described using the Markov Decision Process which is defined by a finite sets of actions and states, transition dynamics which gives us the probability to moving to a next state $s'$ and receiving a reward $r$ after taking action $a$ in state $s$ denoted as $P(s', r | s, a)$ and a reward function giving us the expected reward after executing a certain action in a certain state denoted as $r(s, a) = E[R_t+1 | S_t = s, A_t = a]$ \cite[p.~67]{suttonRLBook}. This setup allows for calculating expected outcomes and planning accordingly with the goal of learning a policy for maximizing the expected reward \cite[p.~67]{suttonRLBook}.

RL algorithms can be roughly divided into three groups: value-based, policy-based, and actor-critic-based methods. Value-based methods learn an optimal value function and derive a policy from it - Q-learning, proposed by \cite{Watkins1992}, is a classic value-based algorithm that learns the optimal action-value function \(Q(s, a)\) by iteratively updating the Q-values based on the Bellman equation. Policy-based methods directly learn a policy without explicitly learning a value function \cite{Bennett2021-qv}. Actor-Critic methods combine aspects of both value-based and policy-based methods - they use an actor to learn a policy and a critic to learn a value function that evaluates the policy \cite{DOGRU20211248}. A description of a candidate method for each group is provided in the Appendix.

\subsection{Large Language models}
LLMs, like the \textit{GPT} family from OpenAI \cite{IntroducingChatGPT_2022}, the \textit{Gemini} family from Google \cite{geminiteam2024geminifamilyhighlycapable}, and the \textit{DeepSeek} family \cite{deepseekai2025deepseekr1incentivizingreasoningcapability}, have dramatically transformed AI, in general, and natural language processing, in particular \cite{minaee2025largelanguagemodelssurvey}. Their ability to perform human-like and even super-human feats in text-based tasks has captured widespread attention \cite{minaee2025largelanguagemodelssurvey, zhao2025surveylargelanguagemodels, Luo_2024}. A key factor driving this success is their extensive pre-training on massive datasets, enabling them to acquire broad world factual knowledge and generate contextually relevant, often insightful, suggestions \cite{roberts2020knowledgepackparameterslanguage, da2021analyzingcommonsenseemergencefewshot}.

These models' ability to generate suggestions and recommendations based on this vast knowledge capture the attention of many scholars \cite{kang2023llmsunderstanduserpreferences, Bao_2023}. \cite{lyu-etal-2024-llm} explored the use of LLMs for personalized recommendation, highlighting their potential to capture user preferences from textual descriptions. Similarly, \cite{kim2024reviewdrivenpersonalizedpreferencereasoning} explores using LLMs to generate personalized product recommendations based on user reviews and product descriptions. \cite{10.1145/3490099.3511105} proposed a system utilizing LLMs to provide creative suggestions for writing and storytelling, demonstrating their ability to generate novel and contextually appropriate ideas.

In a complementary manner, LLMs can generate knowledge, an ability also widely explored and utilized \cite{YANG2025113503, zheng2023largelanguagemodelsscientific}. One can roughly divide the LLM-based knowledge generation into four groups: LLM as a Feature Extractor/Encoder \cite{motger2025leveragingencoderonlylargelanguage}, LLM for Data Augmentation \cite{ding2024dataaugmentationusinglarge}, LLM for Pre-processing/Data Cleaning \cite{zhang2024largelanguagemodelsdata}, and LLM for Post-processing \cite{Wang_2024}. For the first method, \cite{motger2025leveragingencoderonlylargelanguage} utilized the LLM to extract the concrete features of mobile applications being reviewed. Representing the second method, \cite{ding2024dataaugmentationusinglarge} mentioned many different example studies that are using the LLM data augmentation for creating new training data \cite{sahu-etal-2022-data}, data labeling \cite{törnberg2023chatgpt4outperformsexpertscrowd}, data reformation \cite{sharma-etal-2023-paraphrase}, and co-annotations \cite{lu2024doescollaborativehumanlmdialogue}. The third method, exemplified by \cite{zhang2024largelanguagemodelsdata}, utilizes LLM for processing tabular data across four tasks: error detection, data imputation, schema matching, and entity matching, to benchmark different methods and LLMs in this discipline. Finally, for the last method, \cite{Wang_2024} proposed a DiarizationLM framework for post-processing outputs from a speaker diarization system, which could be used for improving the readability of diarized transcripts or reducing the error rate of the word diarization.

Based on this abilities, we adopted three open-source LLMs - DeepSeek-R1 \cite{deepseekai2025deepseekr1incentivizingreasoningcapability}, Llama3.1 \cite{grattafiori2024llama3herdmodels}, and Vicuna \cite{vicuna2023}, to operate as tutors for RL models during the training phase. Notably, while there are performance superior models like GPT and Gemini, the accessibility and open nature of DeepSeek-R1, Llama 3, Vicuna, and other open-source models offer advantages for research and customization while also being more accessible under a tight budget \cite{Shashidhar_2023,hasan2025opensourceaipoweredoptimizationscalene}. A description of each of these models is provided in the Appendix.

\subsection{Integrating Large Language Models with Reinforcement Learning}
The integration of LLMs and RL can be viewed as a bidirectional interaction, with each paradigm offering unique benefits to the other. While the application of RL techniques to LLMs has garnered considerable attention \cite{wang2025reinforcementlearningenhancedllms}, the inverse – leveraging LLMs to enhance RL – remains a comparatively less explored.

The first direction, using RL to improve LLMs, typically involves training LLMs to optimize specific objectives, such as generating more coherent text \cite{kim2025alignstructurealigninglarge, kim2022criticguideddecodingcontrolledtext}, reducing toxicity \cite{Faal_2022, dai2023saferlhfsafereinforcement}, or improving task completion rates \cite{shinn2023reflexionlanguageagentsverbal}. One prominent example is RL from Human Feedback (RLHF) \cite{wang2025reinforcementlearningenhancedllms}, where an RL agent is trained to align LLM outputs with human preferences. In this paradigm, the LLM acts as the policy, and human feedback serves as the reward signal, guiding the LLM to generate outputs that are more desirable from a human perspective. Other examples include using RL to optimize the search process in code generation \cite{le2022coderlmasteringcodegeneration} or to encourage more engaging dialogue in conversational agents \cite{bai2022traininghelpfulharmlessassistant}. These methods have proven highly effective in refining LLM behavior and aligning it with specific human values and goals. Another method is called Direct Preference Optimization (DPO), which abstracts out the need for any reward function by directly using the human preference data - this could lower the complexity of the whole fine-tuning process \cite{wang2025reinforcementlearningenhancedllms}.

In contrast, the second direction – leveraging LLMs to enhance RL – is relatively less explored. As mentioned earlier, LLMs can be used to generate reward functions, interpret policies, or provide structured representations of the environment, all of which can significantly improve the performance and efficiency of RL agents \cite{zhou2024largelanguagemodelpolicy, yang2025leveraging}. For instance, \cite{pang2024illmtscintegrationreinforcementlearning} used an LLM-aided RL agent to improve traffic signal control. The authors developed this strategy due to issues with imperfect communication, including packet loss, delays, or noise, as well as the rare events that could happen in a traffic junction, such as the arrival of emergency vehicles. The authors argue that these factors are often not accounted for in the standard reward functions or training scenarios, thus resulting in non-optimal decisions by the RL agent. 

Subsequently, \cite{Cao_2024} provided a comprehensive survey on the integration of LLMs into RL, highlighting LLMs' ability to leverage reasoning, natural language understanding, and generalization to overcome RL challenges like sample inefficiency, reward design, and generalization in complex environments. Their proposed taxonomy classifies LLM roles in RL into four categories: information processor (extracting features from diverse inputs), reward designer (implicitly or explicitly generating reward functions), generator (providing policy explanations for interpretability), and decision-maker. Crucially, the decision-maker role, which directly addresses sampling and exploration inefficiencies, is further divided into action-making (treating offline RL as a sequence problem leveraging LLM semantic understanding) and action-guiding. In the action-guiding approach, LLMs act as expert instructors, generating candidate or expert actions grounded in their general knowledge, which can then be distilled into an RL agent's policy to enhance sample efficiency. This study specifically adopts this action-guiding framework, directly incorporating LLM-suggested actions based on their general knowledge into the RL agent's policy.

\cite{zhou2024largelanguagemodelpolicy} proposed a novel method called LLM4Teach, which uses LLMs to improve RL training efficiency by training an RL agent with the guidance of an LLM during the early stages of training and gradually transitioning the student to learn from the environment on its own. This strategy allows the RL agent to query the LLM for suggested actions that are delivered as soft, uncertainty-aware instructions. The actions are generated from pre-defined option policies in combination with probability distributions inferred from textual descriptions of the environment the RL agent is solving. Throughout the training, the agent is continuously forced to interact with the environment on its own, thanks to an annealing schedule that lowers the probability of the LLM engagement. Importantly, this study does not consider \say{advice reusing} mechanism \cite{10.5555/3398761.3398953}, which we included in this study.

\cite{yang2025leveraging} proposes a novel method combining imitation learning and LLMs called the Imitation Hierarchical Actor-Critic (IHAC). In general, the method uses a two-phase training strategy: (i) Imitation Learning and (ii) RL. For the first phase, LLM is used to provide high-level strategies that guide the agent's decision-making in the early stages of its training, thus improving its exploration strategy performance by sampling actions from a mixture of the agent's learned policy and LLM-imitated action. Ultimately, the RL agent acquires a strong policy and value function that reflects the general reasoning of the LLM without any further dependence on it. Next, in the second phase, the RL agent continues with the training using a standard actor-critic algorithm method, refining its policy, value functions, etc. Notably, unlike previous methods, phase two does not include LLM at all, making it lightweight and scalable to longer training durations in more complex environments without the worry of additional LLM costs. This method was evaluated in three environments, MiniGrid \cite{10.5555/3666122.3669331}, NetHack \cite{samvelyan2021minihackplanetsandboxopenended}, and Crafter \cite{hafner2022benchmarkingspectrumagentcapabilities}, with the baseline set to results achieved in other papers \cite{zhou2024largelanguagemodelpolicy,prakash2023llmaugmentedhierarchicalagents}, showing it outperforms the baselines in terms of speed, performance, and computational costs for all three cases \cite{yang2025leveraging}.

\subsection{Reusing advice}
In multi-agent RL, the student-teacher framework is a learning paradigm where experienced \say{teacher} agents guide the learning of less experienced "student" agents. This approach's core principle is to leverage the teachers' superior knowledge or experience to accelerate and enhance the learning process of the students within a multi-agent system \cite{DBLP:conf/atal/TorreyT13}. Beyond the initial knowledge transfer in a student-teacher framework, an impactful extension involves simultaneous advising and studying \cite{10.5555/3091125.3091280}. This occurs when a student agent, after effectively learning from one or more teachers, subsequently becomes a teacher itself for other student agents. This cascading knowledge transfer offers significant advantages, especially in large-scale multi-agent reinforcement learning (RL) systems. First, it enables the efficient dissemination of expertise throughout the agent population. Second, it promotes peer-to-peer knowledge propagation, reducing the reliance on and burden of a single expert teacher, thereby enhancing scalability. Finally, this iterative learning process can foster more robust and adaptable learning. Students exposed to multiple "generations" of teachers may develop a more nuanced understanding of the environment and become less prone to overfitting to the behaviors of any single teacher.

\cite{10.5555/3398761.3398953} introduced reusing of the given advice in RL by proposing that if an agent receives advice in a given state, the state will be labeled as \say{advised}, and the state-advice pair will be stored in memory for later usage. Next time the RL agent visits the same state, it might reuse the advice based on some probability function \cite{10.5555/3398761.3398953}. Despite this method's potential, if the student model insists on some non-optimal advice for too long, it might hinder their overall performance \cite{10.5555/3398761.3398953}. The student should instead intelligently choose a proper time to reuse the advice \cite{10.5555/3398761.3398953}. Thus, the authors propose three approaches to accomplish this: (i) Q-change per Step - the advice will be reused if the suggested action leads to a higher Q value; (ii) ReBudget - insisting on given advice for some time is likely to stabilize the performance, this approach will apply the advice in the given step N times; and (iii) Decay - each time an agent visits an advised state, it will use the suggestion with a given probability which will decay every time the advice is used \cite{10.5555/3398761.3398953}. In this study, we adopted the second option due to its straightforward manipulation for various experimental settings.

\section{Methods and Materials}
\label{sec:methods}

\subsection{Student-teacher architecture for LLM-RL with repeated advice mechanism}\label{studentTeacher}
\begin{figure}[!htbp]
    \centering
    \makebox[\textwidth][c]{\includegraphics[width=1\linewidth]{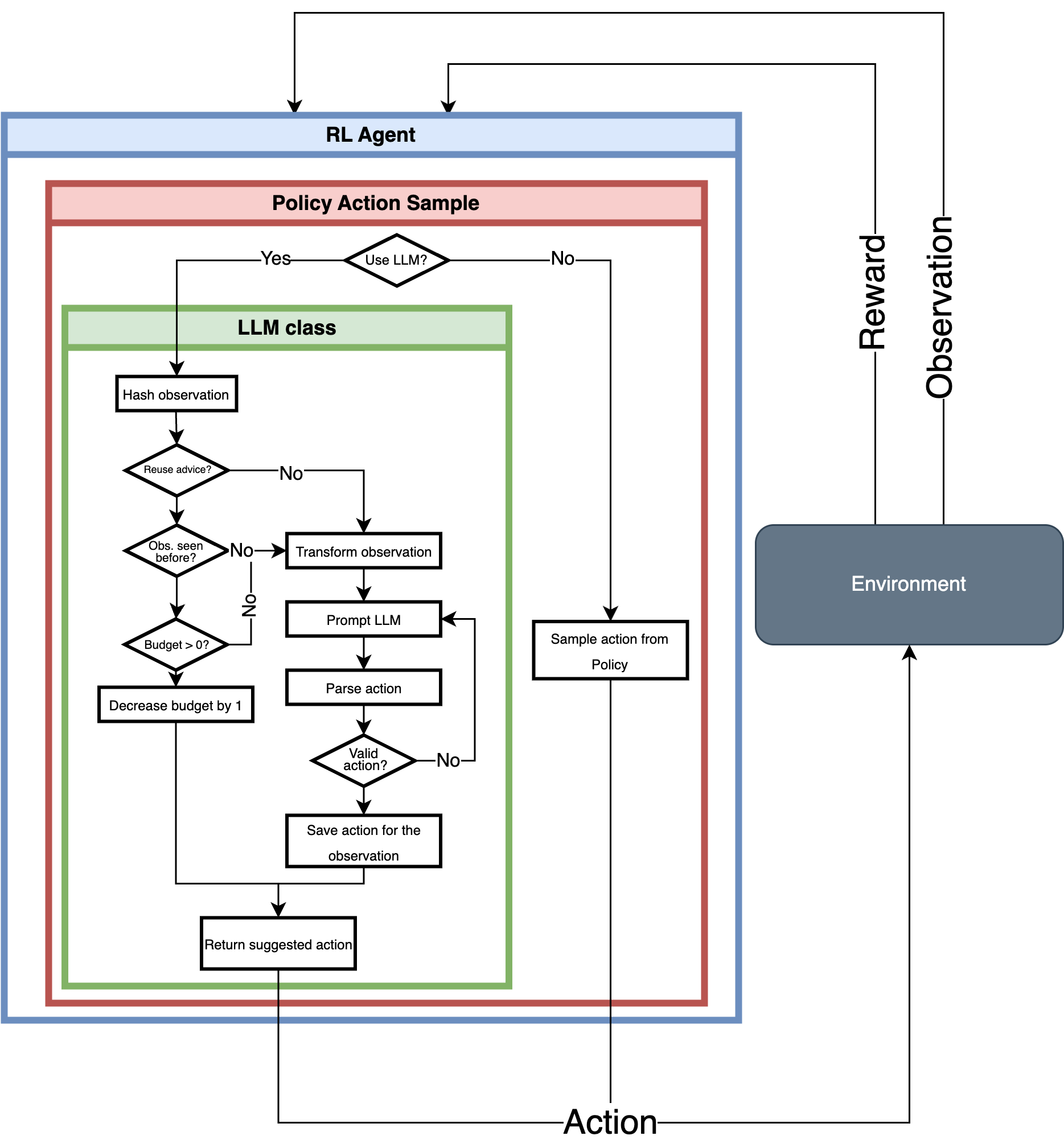}}
    \caption{A schematic view of student-teacher structure for RL and LLM. When the RL agent selects an action to execute in the environment, it has the opportunity to consult the LLM tutor. The decision to do so is guided by a probability \textbf{P}, continuously decaying during the training phase.}
    \label{fig:implementation}
\end{figure}

In this section, we outline the proposed student-teacher architecture for LLM-RL with the advice reusing mechanism. Fig. \ref{fig:implementation} presents a schematic view of the proposed student-teacher structure for RL and LLM. When the RL agent needs to select an action to execute in the environment, it has the opportunity to consult the LLM tutor. The decision to do so is guided by a probability \textbf{P}, continuously decaying during the training phase.

\subsubsection{LLM Implementation}\label{llmClass}
Algorithm \ref{setActionPseudoCode} outlines the RL-LLM as student-teacher with the reusing mechanism. The algorithm accepts a state \(s'\) and returns an LLM-advised action based on the given state. First, a hash of the state is computed using a hashing function (line 2). If this hash exists in a look-up dictionary $O$ (line 3), the corresponding action object $a$ is retrieved (line 4). If reusing actions is configured for the experiment running this function and the object still has a budget greater than zero (line 5), the budget is decremented (line 6) and the stored action is returned (line 7). If no valid reusable action exists, the function enters a loop (line 10) that continues until a valid and applicable suggestion $g$ is generated by an LLM (line 11). Once such advice is found, a new action and a default budget of 3 are assigned (line 13) to a new object $a$, and this object is stored in the dictionary using the computed hash (line 14). Finally, the newly generated action is returned (line 15).

To ensure the reliable extraction of advice, the LLM is programmatically instructed to format its suggestions within <action></action> tags. However, several potential issues can arise at this stage: the system may fail to parse the action from the LLM's response, the suggested action may not conform to the expected integer format, or it might prove inapplicable within the given environment. In the former two scenarios, the system defaults to a randomly selected action. In the latter, an invalid action might initially be accepted, with its incorrectness only becoming apparent at a later stage of execution. In any of these error conditions, the LLM is re-prompted until a valid action is successfully acquired. Once validated, this action is then archived alongside its corresponding state for future reference.

\begin{algorithm}
\caption{Set Action Method}
\label{setActionPseudoCode}
\begin{algorithmic}[1]
    \STATE \textbf{Function} \textsc{set\_action}($s'$)
    \STATE $hash \gets \textsc{hashing\_function}(s')$
    \IF{$hash \in O$}
        \STATE $a \gets O[hash]$
        \IF{$reusing\_allowed \wedge a.budget > 0$}
            \STATE $a.budget \gets a.budget - 1$
            \RETURN $a.action$
        \ENDIF
    \ENDIF
    \WHILE{$g$ is not valid \textbf{and} $g$ is not applicable}
        \STATE $g \gets \textsc{generate\_llm\_advice}(s')$
    \ENDWHILE
    \STATE $a.action, a.budget \gets g, 3$
    \STATE $O[hash] \gets a$
    \RETURN $a.action$
    \STATE \textbf{End Function}
\end{algorithmic}
\end{algorithm}

\subsubsection{Altering the policy}\label{sec:alteringThePolicy}
In order to enable the LLM to influence the RL agent, it needs to be integrated directly into the agent's training process. For instance, in a Deep Q-Network (DQN) setup, the traditional random exploration strategy is replaced with a probabilistic method. At each decision point, the agent can either follow its learned policy or seek guidance from the LLM, as adapted from \cite{zhou2024largelanguagemodelpolicy}. A probability determines whether the action is chosen by the agent's policy or obtained as advice from the LLM. If the LLM is selected, it is queried for a recommended action. For Proximal Policy Optimization (PPO) and Advantage Actor-Critic (A2C) algorithms, a similar integration approach was adopted. The \texttt{forward} method of the actor-critic policy, which dictates action selection, was adapted to probabilistically choose between actions from the learned policy and those advised by the Large Language Model (LLM), mirroring the strategy used for Deep Q-Network (DQN). The architectural demands of actor-critic algorithms necessitated further adjustments to maintain training compatibility. Specifically, the \texttt{value} and \texttt{log\_prob} outputs, essential for estimating state values and computing policy gradients, were similarly generated even when an LLM-suggested action was chosen. 

\subsubsection{Budgets and probabilities}
When an action is sampled from the policy, a pseudo-random number between 0 and 1 is generated and compared against the probability of querying the LLM for guidance. This mechanism is inspired by previous work (e.g., \cite{intro_28}), which utilized a decaying probability for LLM tutor engagement, starting at 1.0 and decreasing to 0.1 over a set number of training iterations. Following this, the probability was either set to 0 or maintained at 0.1 for the remainder of the training. Our approach adopts the same decay strategy, with the probability linearly decreasing from 1.0 to a final value of 0.1, as defined in Eq. (\ref{eq:updated_probability}):

\begin{equation}
    P_{updated} = P_{initial} - \frac{\tau}{\theta}(P_{initial} - P_{final}) ,
    \label{eq:updated_probability}
\end{equation}
where $\tau \in \mathbb{N}$ is the number of steps taken by the agent, and $\theta  \in \mathbb{N}$ represents the total number of steps the agent can take before the probability of using the LLM reaches its final value.

This adjustment is integrated directly into the RL agent's training process, gradually reducing LLM reliance over time. Notably, the budget for LLM-suggested actions was set to 3, consistent with the methodology in \cite{10.5555/3398761.3398953}. Once the probability of consulting the LLM reaches 0.1, it remains constant for the duration of the training period.

\subsubsection{Prompt engineering}
To make sure the LLM-suggested actions are reliable, it is important to tell the model what it should consider. To this end, the observations are transformed before prompting the LLM. This transformation is essential to ensure that the LLM understands the task and can provide meaningful suggestions. Each environment used during the experiments has its own transform function, which accepts a raw environment observation and outputs a natural language prompt for the LLM. Instruction prompting \cite{sanh2022multitaskpromptedtrainingenables} was used to directly tell the LLM to follow the instructions, which should enable broader generalization. In addition, Zero-shot prompting \cite{brown2020languagemodelsfewshotlearners} was also utilized, as we did not give the model any examples of how to solve the problem. Hence, we expected the model to generalize purely from the instructions and data. Lastly, Structured Output Format\cite{pang2024illmtscintegrationreinforcementlearning,chen2023programthoughtspromptingdisentangling,zhong2020semanticevaluationtexttosqldistilled} is used to tell the LLM how we expect the output to look.  Moreover, a model file serves as a blueprint for an LLM, allowing Ollama to create a custom model based on an existing one \cite{Marcondes2025}. In this study, model files were used to define the system prompts for all employed LLMs. The system prompt determines the system message that is inserted into the template \cite{zhang2024sprigimprovinglargelanguage}.

For instance, listing \ref{transformObservation} shows an example of how the raw observation from the Snake environment is transformed into a natural language prompt before being provided to the LLM. Namely, the game’s state is described using both a 2D grid and plain-language explanations. The grid shows the full environment layout, where the snake’s head, body, and food are represented by numerical values. Additional context is provided to highlight the snake’s head and the food's exact positions, helping the model better interpret the situation. This prompt uses instruction prompting by clearly telling the LLM what it should do—namely, to analyze the current state and suggest the next action. Zero-shot prompting is applied here as well since the model is expected to respond correctly without being shown any prior examples. Lastly, the response format is controlled using a structured output format, where the expected action must be placed within <action></action> tags. This structure makes it easier for the system to extract and use the LLM’s response during training.

\begin{lstlisting}[caption={Transformed observation for the Snake environemnt}, label=transformObservation]
This is the current 2D grid of the snake game:
[[0 0 0 0 0 0 0 0 0 0]
[0 0 0 0 1 0 0 0 0 0]
[0 0 0 0 0 0 0 0 0 0]
[0 0 0 0 0 0 0 0 0 0]
[ 0  0  0  0  0 -1  0  0  0  0]
[ 0  0  0 -1 -1 -1  0  0  0  0]
[0 0 0 0 0 0 0 0 0 0]
[0 0 0 0 0 0 0 0 0 0]
[0 0 0 0 0 0 0 0 0 0]
[0 0 0 0 0 0 0 0 0 0]]
The snake's head is located at row 4, column 5, and the food
is located at row 1, column 4. The rest of the snake's body
is represented by -1.
Please clarify the current state of the game and determine what
the agent should do in this current state. 
Finally, please suggest the correct action and
output its index in the <action></action> tags. 
Do not provide reasoning.
\end{lstlisting}

Furthermore, listing \ref{systemMessageTemplate} displays a model file template. We used this template for all LLMs in this study. This template includes several prompt engineering concepts:Instruction and Structured Output prompting, Role Prompting \cite{white2023promptpatterncatalogenhance}, Contextual Grounding  \cite{brown2020languagemodelsfewshotlearners}, and Constrained Action Prompting \cite{yao2023reactsynergizingreasoningacting}, to ensure a robust and relevant response from the LLMs.

\begin{lstlisting}[caption={Template of a system message}, label=systemMessageTemplate]
You are a system used as a teacher for Reinforcement
learning (RL) agent. Your goal is to use your reasoning
to help with the convergence of optimal policy of
the RL agent.
The environment you will guide the agent is <name of the game>
You will be given the <observation space description>.

You can select an action from this dictionary: <dictionary 
of action mapping integer values to a string representation
of an action>.

Output: Clarify the current state and suggest the best
action for the agent to take. Output the action's
index (key from the actions dictionary) in the
<action></action> tags (e.g. <action>3</action>).
\end{lstlisting}

\subsection{Experimental setup}
This section outlines the experiments proposed to evaluate the implemented solution. These experiments can be categorized into three groups based on the environment: Blackjack, Connect Four, and Snake. Within each category, DQN, PPO, and A2C were executed both without any LLM (as the baseline) and with three different LLMs: Llama3.1, Vicuna, and DeepSeek R1. These experiments aimed to leverage the selected LLMs to assist the learning of an RL agent as teachers and to observe whether this support helps the agent converge to the optimal solution faster than in the absence of an LLM teacher. The environments are described sequentially, followed by the selected RL algorithms, and finally the chosen LLM models. 

All of the Vicuna and LLaMA experiments were run on the Jönköping University cluster with an NVIDIA A100 GPU with 80GB of memory. Due to restrictions on the server, each student can use just one of seven MIG \cite{vamja2025partitioninggpupowermultiinstances} slices, restricting the total GPU memory available during experiments to just 10 GB. The DeepSeek-R1 experiments were run on a MacBook Pro with an M4 CPU and 24 GB of RAM.

\subsubsection{Environments}
We used three popular games as the RL environments. Starting with the Blackjack game, a simplified digital version of the classic card game, where the player competes against a dealer, is used. The player’s score and visible cards are shown on the screen. For the Connect Four game, a two-player board game environment where agents take turns dropping colored tokens into a vertical grid, aiming to connect four of their own pieces in a row, column, or diagonal is adopted. Lastly, for the Snake game, we used the classic arcade-style game where the player controls a moving snake that must collect food while avoiding collisions with its own body and the edge of the game's map. These environments are chosen to capture a diverse set of tasks, allowing a more robust evaluation of the integration of LLM-guided decision-making in RL. Fig. \ref{fig:environments} illustrates the three environments.

\begin{figure}
\begin{tabular}{ccc}
  \includegraphics[width=45mm,height=45mm]{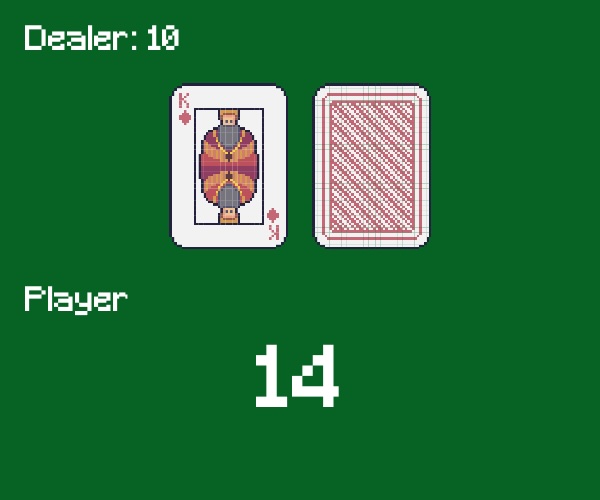} &   \includegraphics[width=45mm,height=45mm]{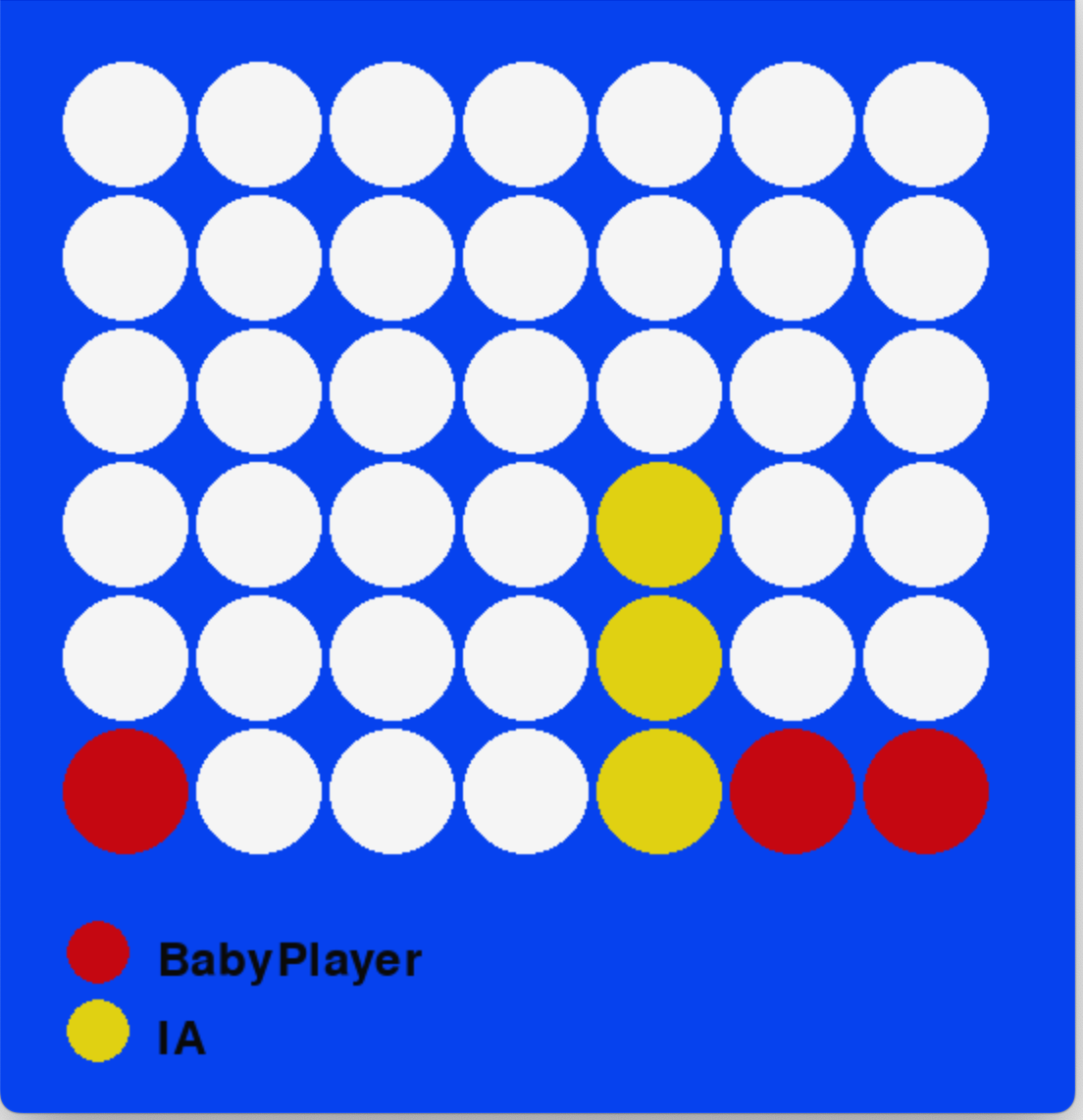} & \includegraphics[width=45mm,height=45mm]{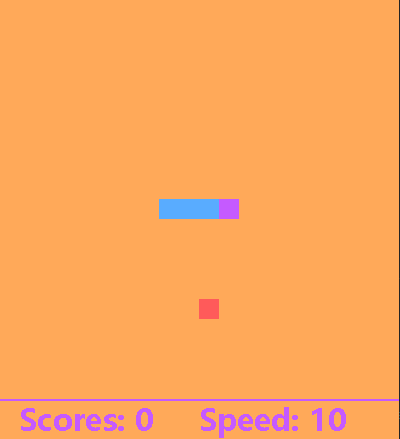} \\
(i) Blackjack & (ii) Connect Four  & (iii) Snake \\[6pt]
\end{tabular}
\caption{Visual presentation of all three selected RL environments.}
\label{fig:environments}
\end{figure}

To ensure consistency and comparability across all three environments, we adopted the Gymnasium framework \cite{towers2024gymnasiumstandardinterfacereinforcement}. This framework provides a standardized interface for constructing RL environments and managing their interactions with learning agents. A formal description of each environment is provided in the Appendix.

\subsubsection{RL algorithms}\label{rlAlgorithms}
We evaluated three distinct RL algorithms: DQN, PPO, and A2C, in order to ensure a comprehensive exploration of the solution space, given their unique algorithmic foundations. Specifically, DQN was chosen as a representative off-policy RL algorithm. In contrast, both PPO and A2C are on-policy, actor-critic algorithms, which fundamentally alter how an agent learns its optimal policy. The specific hyperparameter values used for each of these RL algorithms are detailed in Table \ref{tab:hyperparameters}.

\begin{table}[ht]
    \centering
    \caption{Hyperparameters used for each RL algorithm}
    \label{tab:hyperparameters}
    \begin{tabular}{lccc}
         \hline \hline
        \textbf{Hyperparameter} & \textbf{DQN} & \textbf{PPO} & \textbf{A2C} \\
         \hline \hline
        Learning Rate           & 0.0001  & 0.0003 & 0.0007         \\
        Gamma (Discount Factor) & 0.99   & 0.99   & 0.99             \\
        Optimizer & Adam   & Adam   & Adam           \\
        Clip Range              & -      & 0.2    & -           \\
        Memory Buffer Size      & 1000     & -     & -            \\
        Batch Size              & 32     & 64     & 5             \\
        \hline \hline
    \end{tabular}
\end{table}

\subsubsection{Pre-trained LLMs}\label{sec:pretrainedLLMs}
Local LLMs were selected for two key reasons. First, the high cost associated with cloud-based LLMs makes them less practical for extended experimentation, especially under limited resources. Second, there is a growing interest in exploring the potential of locally hosted LLMs, which are rapidly gaining popularity due to their increasing performance, accessibility, and privacy \cite{bumgardner2023locallargelanguagemodels, Grothey2025, Wiest2024}. To this end, we adopted the following three LLM models: DeepSeek-R1 (14b parameters), Vicuna (13b parameters), and LLaMA3.1 (8b parameters). These three models can be run on relatively modest computational settings and are open-source. 

The number of iterations over which the probability of consulting the LLM teacher decayed to its final value of 0.1 varied across environments. For the Blackjack environment, the decay occurred over 3000 steps within a total of 15000 training steps. In the Connect Four environment, 1000 steps were allocated for probability decay out of 10000 total steps per experiment. Finally, for the Snake environment, 1000 steps were dedicated to decaying the LLM usage probability within a total of 8000 training steps. These numbers were picked following a manual trial-and-error process.

\subsection{Relative RL Convergence Metric}\label{sec:convergence_metric}
In order to measure the performance of the RL's convergence, we define a metric that captures the RL's performance over the training process. Formally, let us represent the performance over the training of an RL algorithm as a vector, \(v\), where the index of the vector is the training step and the corresponding value is the model's performance with respect to the reward function used to train the RL algorithm. In order to obtain a comparable definition of a set of vectors \(V\) that represent a set of RL algorithm training sessions with the same reward function but different configurations, we first normalize these vectors. Formally, we divide the indexes of each \(v \in V\) by the size of \(v\) and linearly scaling the performance of the vector with respect to all vectors - i.e., \(\forall v \in V, \forall p \in \hat{p} := (p-min(P)/(max(P) - min(P))\) where \(min(P) := \min_{p \in v, v\in V} p\) and  \(max(P) := \max_{p \in v, v\in V} p\). The scaling uses global $min$ and $max$ to ensure that all vectors are normalized on the same scale. Hence, the RL's performance during training is defined to be:
\begin{equation}
    P := \int_{0}^{1} \hat{p}(t) dt.
    \label{eq:main}
\end{equation}

\section{Results}
\label{sec:results}
A total of 54 experiments were conducted in various RL environments. Fig. \ref{fig:environments} provides a comparative overview of all experiments conducted across the Blackjack, Connect Four, and Snake environments, utilizing the DQN, PPO, and A2C algorithms. The plots are arranged in a grid, with each row representing an algorithm and each column a distinct environment. Each subfigure displays the respective learning curves, offering an immediate visualization of performance evolution during training. To enhance clarity for the Blackjack environment, the figures are smoothed using a multivariate spline algorithm \cite{2d4bd73b-c6a2-3e01-827a-0fb131aca4a0}  with a smoothing factor of 7. The impact of the LLM tutor is most pronounced in the Connect Four and Snake environments. Generally, the smallest LLM, LLaMA, tended to underperform compared to the baselines. In contrast, the larger LLMs, Vicuna and DeepSeek, yielded superior performance, either matching or surpassing the baselines.

\begin{figure}[htbp]
\centering
\begin{subfigure}[t]{0.3\textwidth}
  \includegraphics[width=\linewidth]{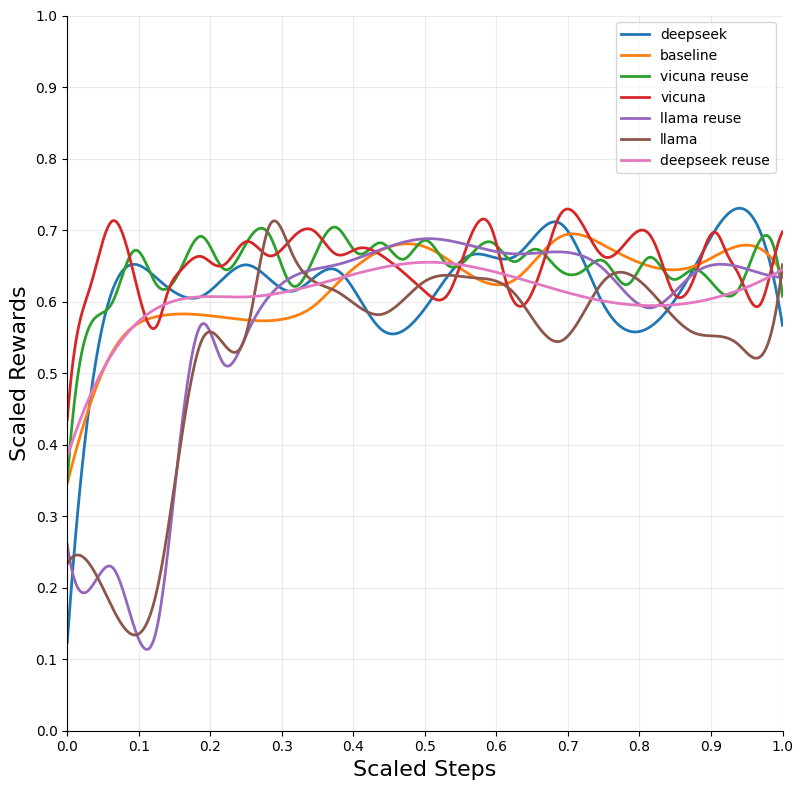}
  \caption{Blackjack — DQN}
  \label{fig:blackjack_dqn}
\end{subfigure}
\hfill
\begin{subfigure}[t]{0.3\textwidth}
  \includegraphics[width=\linewidth]{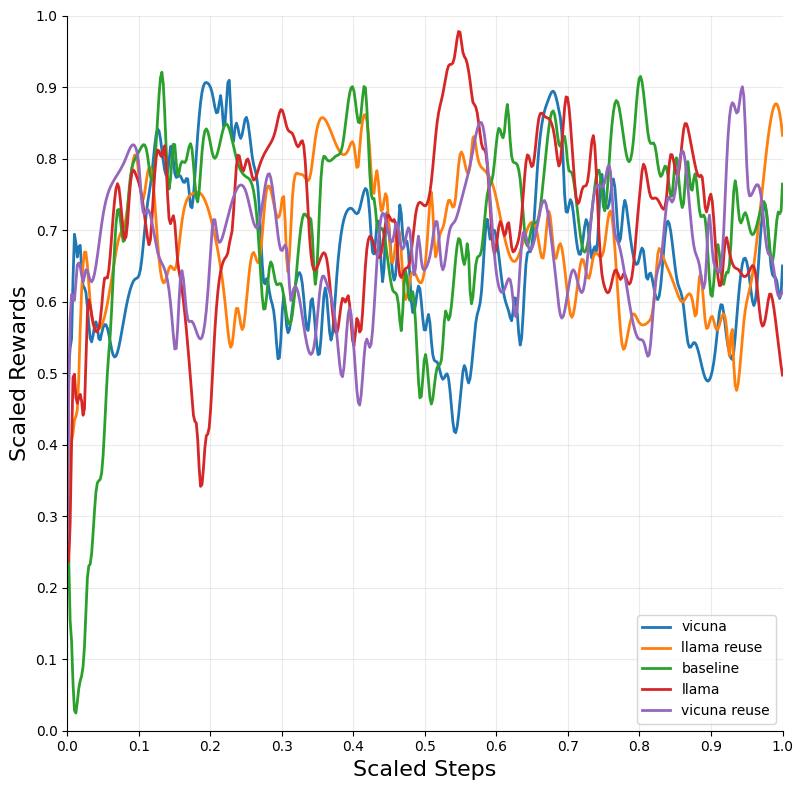}
  \caption{Connect Four — DQN}
  \label{fig:c4_dqn}
\end{subfigure}
\hfill
\begin{subfigure}[t]{0.3\textwidth}
  \includegraphics[width=\linewidth]{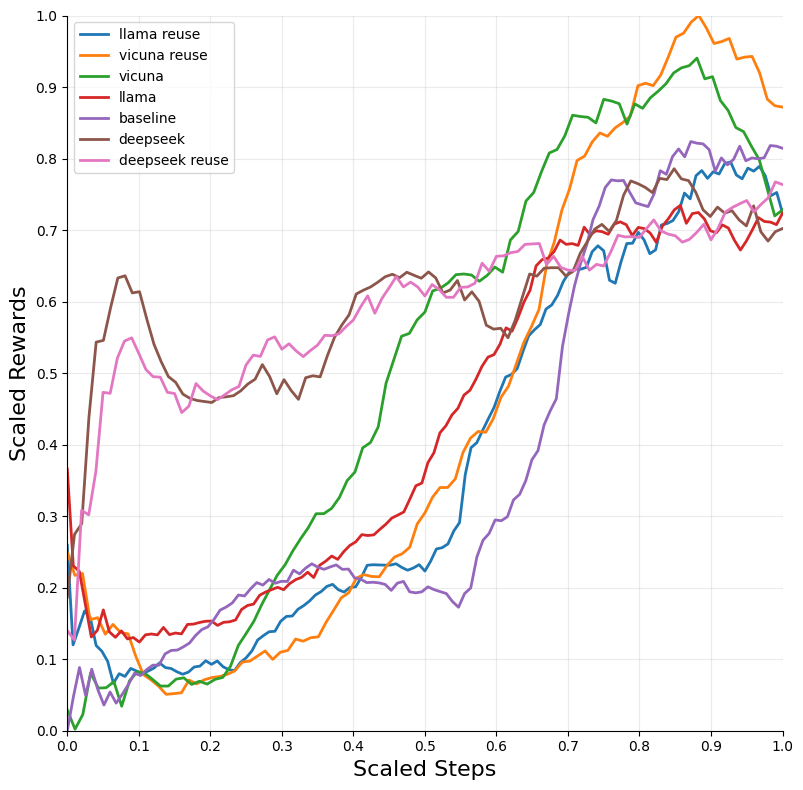}
  \caption{Snake — DQN}
  \label{fig:snake_dqn}
\end{subfigure}

\vspace{1em}

\begin{subfigure}[t]{0.3\textwidth}
  \includegraphics[width=\linewidth]{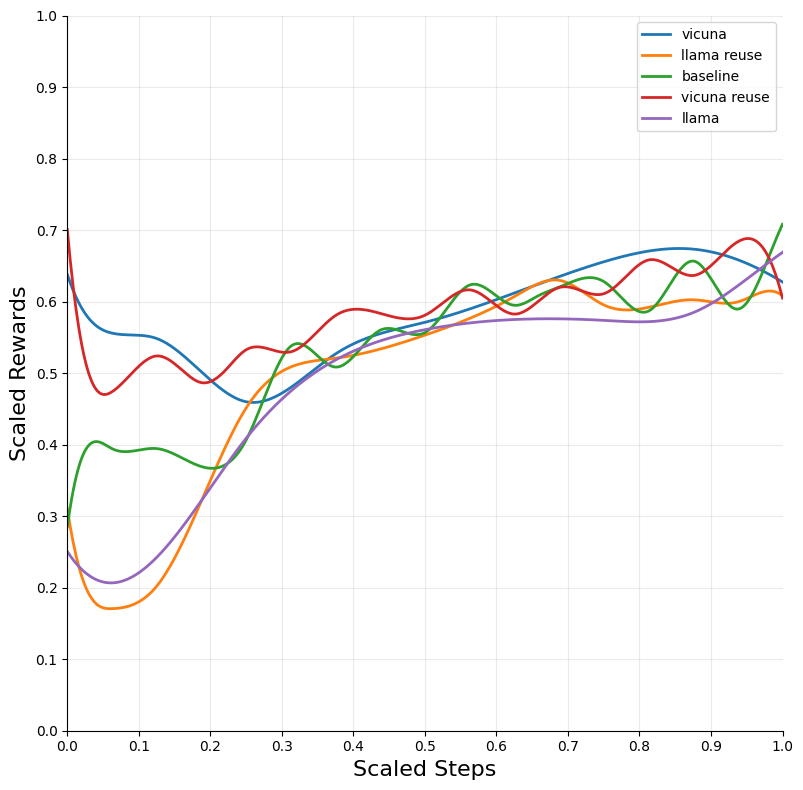}
  \caption{Blackjack — PPO}
  \label{fig:blackjack_ppo}
\end{subfigure}
\hfill
\begin{subfigure}[t]{0.3\textwidth}
  \includegraphics[width=\linewidth]{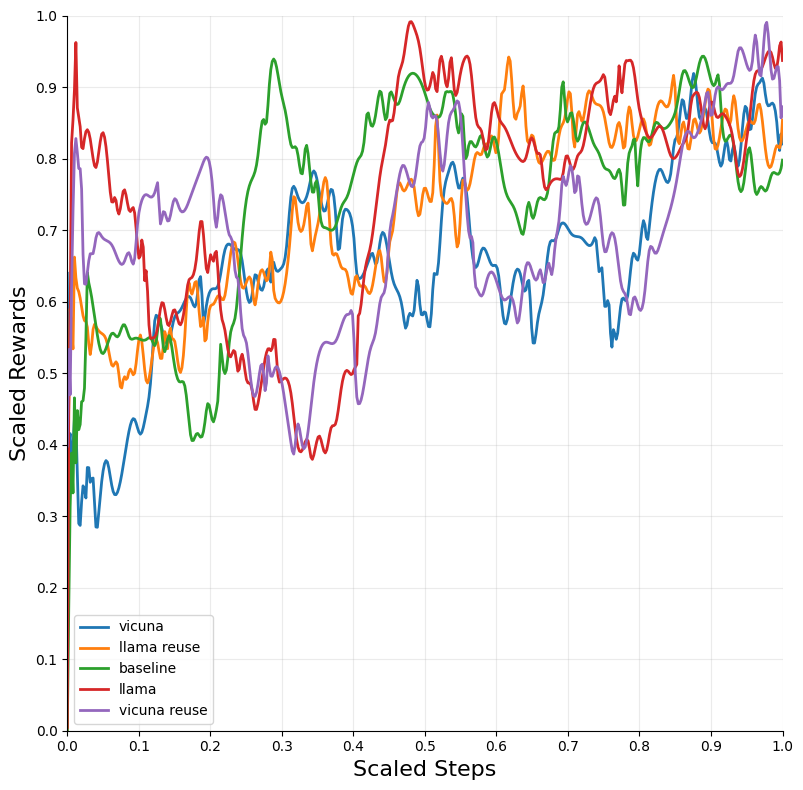}
  \caption{Connect Four — PPO}
  \label{fig:c4_ppo}
\end{subfigure}
\hfill
\begin{subfigure}[t]{0.3\textwidth}
  \includegraphics[width=\linewidth]{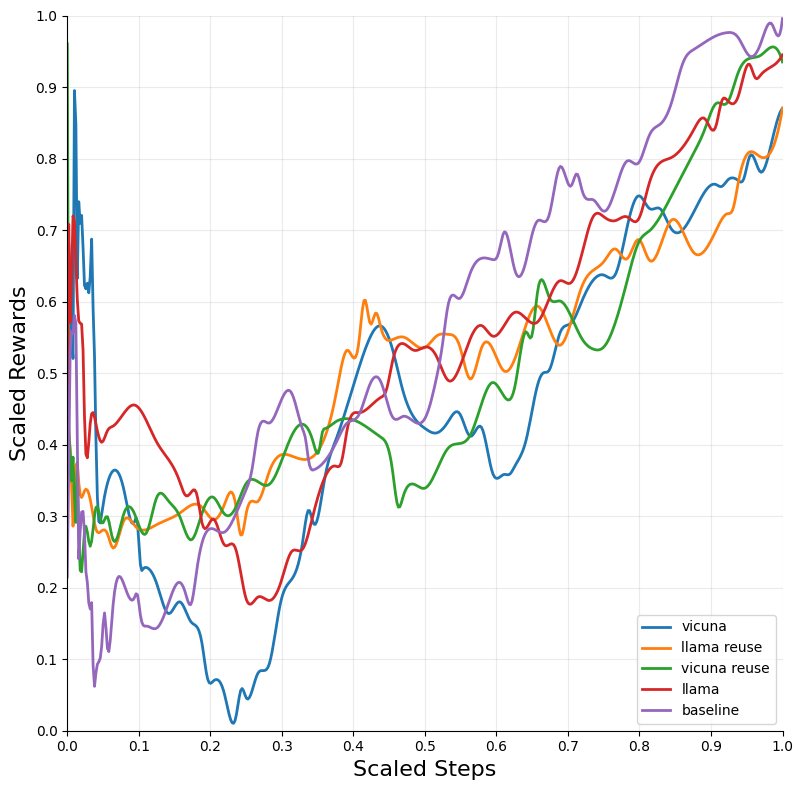}
  \caption{Snake — PPO}
  \label{fig:snake_ppo}
\end{subfigure}

\vspace{1em}

\begin{subfigure}[t]{0.3\textwidth}
  \includegraphics[width=\linewidth]{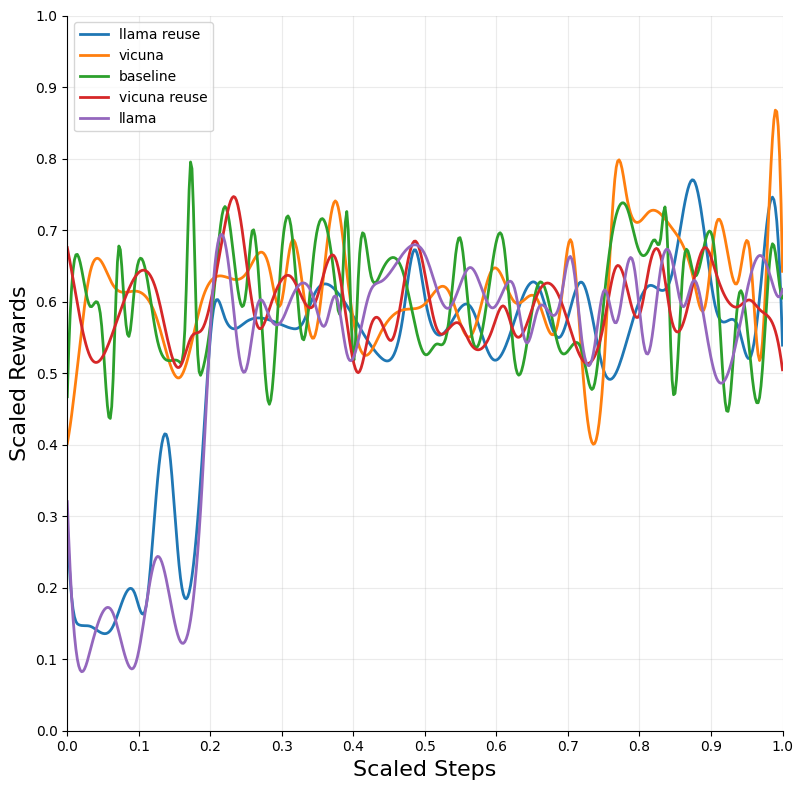}
  \caption{Blackjack — A2C}
  \label{fig:blackjack_a2c}
\end{subfigure}
\hfill
\begin{subfigure}[t]{0.3\textwidth}
  \includegraphics[width=\linewidth]{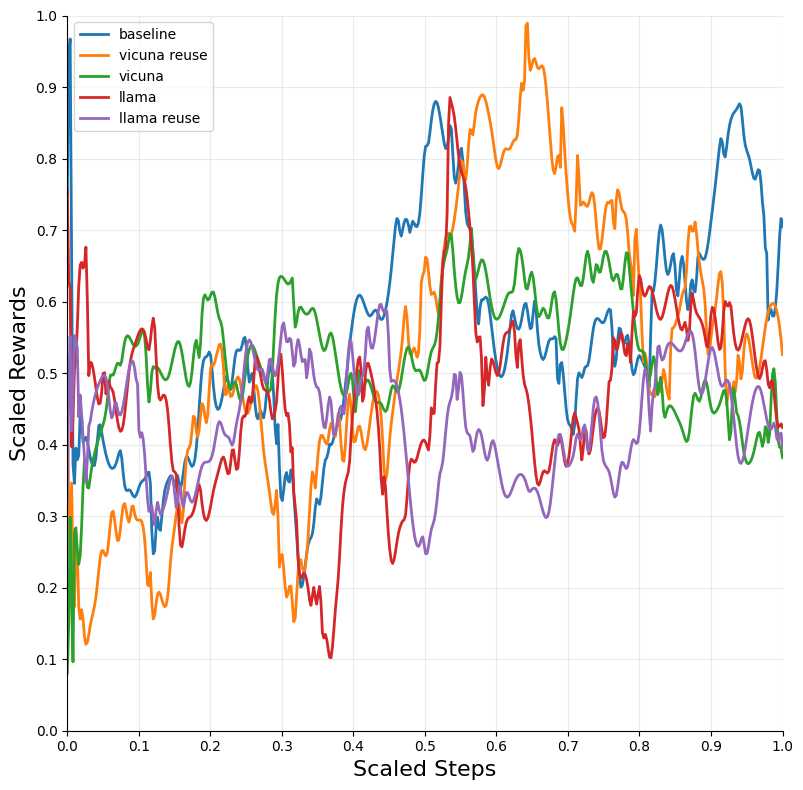}
  \caption{Connect Four — A2C}
  \label{fig:c4_a2c}
\end{subfigure}
\hfill
\begin{subfigure}[t]{0.3\textwidth}
  \includegraphics[width=\linewidth]{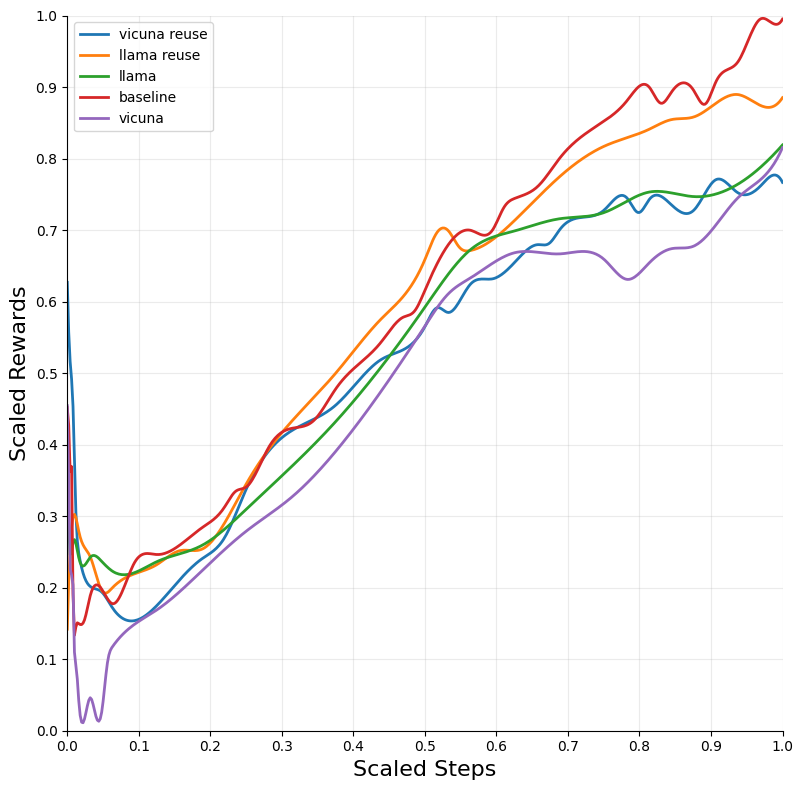}
  \caption{Snake — A2C}
  \label{fig:snake_a2c}
\end{subfigure}

\caption{The normalized collected reward over training for all three RL algorithms with their respective LLM teachers, divided into the three environments. }
\label{fig:environments}
\end{figure}

Table \ref{table:main_result} presents the performance (\(P\)) of the three RL algorithms with the three LLM models (and a baseline) across three environments. Notably, on average across all environments, the DeepSeek-enhanced models outperform the baselines (0.6058 vs 0.5811), Vicuna performs on par (0.5798 vs 0.5811), while LLaMA trails slightly behind (0.5425 vs 0.5811). The DQN-based models exhibited the strongest effect sizes, suggesting that LLM augmentation provides substantial performance gains in this setting. Moreover, the \say{reuse} variants, which leverage past interactions with the LLM, tend to yield even higher scores. From these results, it seems there is a directly proportional relationship between the LLM size and the performance of the proposed methods. To empirically establish this connection, we computed the Pearson correlation between the LLM size and observed performance \cite{Sedgwicke4483}, obtaining that the size of the LLM correlates with the final result, with a coefficient of 0.24 and a p-value of 0.14.

\begin{table}[!ht]
    \centering
    \begin{tabular}{ccccc}
         \hline \hline
        \textbf{Environment} & \textbf{RL algorithm} & \textbf{LLM model} & \textbf{Reuse} & \textbf{Performance} \\
        \hline \hline
        \multirow{15}{*}{Blackjack} 
            & \multirow{5}{*}{PPO} 
                & Vicuna & No & 0.5821 \\
                &  & Vicuna & Yes & 0.5822 \\  \cline{3-5}
                &  & LLaMA & No & 0.4874 \\
                &  & LLaMA & Yes & 0.4932 \\ \cline{3-5}
                &  & Baseline & - & 0.5345 \\
            \cline{2-5}
            & \multirow{5}{*}{A2C} 
                & Vicuna & No & 0.6175 \\
                &  & Vicuna & Yes & 0.5938 \\ \cline{3-5}
                &  & LLaMA & No & 0.5152 \\
                &  & LLaMA & Yes & 0.5173 \\ \cline{3-5}
                &  & Baseline & - & 0.6052 \\
            \cline{2-5}
            & \multirow{7}{*}{DQN} 
                & Vicuna & No & 0.6549 \\
                &  & Vicuna & Yes & 0.6494 \\ \cline{3-5}
                &  & LLaMA & No & 0.5328 \\
                &  & LLaMA & Yes & 0.5671 \\ \cline{3-5}
                &  & DeepSeek & No & 0.6211 \\
                &  & DeepSeek & Yes & 0.6073 \\ \cline{3-5} 
                &  & Baseline & - & 0.6216 \\
        \hline
        \multirow{15}{*}{Connect Four} 
            & \multirow{5}{*}{PPO} 
                & Vicuna & No & 0.7259 \\
                &  & Vicuna & Yes & 0.7451 \\ \cline{3-5}
                &  & LLaMA & No & 0.6550 \\
                &  & LLaMA & Yes & 0.7455 \\ \cline{3-5}
                &  & Baseline & - & 0.6938 \\
            \cline{2-5}
            & \multirow{5}{*}{A2C} 
                & Vicuna & No & 0.5252 \\
                &  & Vicuna & Yes & 0.5235 \\ \cline{3-5}
                &  & LLaMA & No & 0.4648 \\
                &  & LLaMA & Yes & 0.4326 \\ \cline{3-5}
                &  & Baseline & - & 0.5454 \\
            \cline{2-5}
            & \multirow{5}{*}{DQN} 
                & Vicuna & No & 0.6620 \\
                &  & Vicuna & Yes & 0.6754 \\ \cline{3-5}
                &  & LLaMA & No & 0.7080 \\
                &  & LLaMA & Yes & 0.6824 \\ \cline{3-5}
                &  & Baseline & - & 0.7066 \\
        \hline
        \multirow{15}{*}{Snake} 
            & \multirow{5}{*}{PPO} 
                & Vicuna & No & 0.4551 \\
                &  & Vicuna & Yes & 0.4958 \\ \cline{3-5}
                &  & LLaMA & No & 0.5402 \\
                &  & LLaMA & Yes & 0.5092 \\ \cline{3-5}
                &  & Baseline & - & 0.5557 \\
            \cline{2-5}
            & \multirow{5}{*}{A2C} 
                & Vicuna & No & 0.4792 \\
                &  & Vicuna & Yes & 0.5226 \\ \cline{3-5}
                &  & LLaMA & No & 0.5333 \\
                &  & LLaMA & Yes & 0.5842 \\ \cline{3-5}
                &  & Baseline & - & 0.5967 \\
            \cline{2-5}
            & \multirow{7}{*}{DQN} 
                & Vicuna & No & 0.5005 \\
                &  & Vicuna & Yes & 0.4379 \\ \cline{3-5}
                &  & LLaMA & No & 0.4192 \\
                &  & LLaMA & Yes & 0.3768 \\ \cline{3-5}
                &  & DeepSeek & No & 0.6023 \\
                &  & DeepSeek & Yes & 0.5926 \\  \cline{3-5}
                &  & Baseline & - & 0.3702 \\
        \hline \hline
    \end{tabular}
    \caption{The normalized performance over training session for each combination of environment, RL algorithm, and LLM model.}
    \label{table:main_result}
\end{table}

Table \ref{table:timeSave} presents an overview of the time saved in minutes during the training of RL agents when reusing advice generated by the selected LLMs across the three selected environments: Blackjack, Connect Four, and Snake. The table includes the average number of advice reuses, the LLM's average response time in seconds, and the corresponding training time saved. Notably, DeepSeek consistently results in the highest time savings across all environments (450 minutes in Blackjack, 596 minutes in Connect Four, and 46 minutes in Snake) due to the number of tokens it generates while providing advice to the agent. In contrast, LLaMA delivers the fastest responses across all environments. Vicuna shows moderate response times and time savings. 

\begin{table}[!ht]
    \centering
    \resizebox{\textwidth}{!}{
    \begin{tabular}{ccccc}
         \hline \hline
        \textbf{Environment} & \textbf{Average number of reuses} & \textbf{LLM} & \textbf{LLM Average response time (Seconds)} & \textbf{Saved Time (Minutes)} \\
        \hline \hline
        \multirow{3}{*}{Blackjack} 
            & \multirow{3}{*}{901} & Vicuna & 7.33 & 110\\
            &                         & LLaMA & 2.48 & 37 \\
            &                         & DeepSeek & 29.94 & 450 \\
        \hline
        \multirow{3}{*}{Connect Four} 
            & \multirow{3}{*}{172} & Vicuna & 8.52 & 24\\
            &                         & LLaMA & 3.33 & 10 \\
            &                         & DeepSeek & 208 & 596 \\
        \hline
        \multirow{3}{*}{Snake} 
            & \multirow{3}{*}{97} & Vicuna & 7.12 & 12\\
            &                         & LLaMA & 3.36 & 5 \\
            &                         & DeepSeek & 28.33 & 46 \\
        \hline \hline
    \end{tabular}
    }
    \caption{The time saved during training due to reusing advice per environment, RL algorithm, and LLM model.}
    \label{table:timeSave}
\end{table}

\section{Discussion}
\label{sec:discussion}
In this study, we explore whether advice from LLM tutors could expedite convergence toward optimal performance for RL algorithms and also improve their overall performance. In particular, we investigated the influence of the \say{advise reuse} mechanism in this context. To this end, we conducted a robust exploration of these two hypotheses by exploring three RL algorithms as students (DQN, PPO, and A2C) together with three pre-trained LLM models (Llama, Vlcuna, and DeepSeek) as tutors on three environments (the Blackjack game, the Snake game, and the Connect Four game). For each combination of these three, we computed the convergence (see Eq. (\ref{eq:main})) between the LLM-tutored RL model and the baseline (without LLM tutoring) and the advice reusing mechanism-enhanced version. 

Our results clearly indicate that pre-trained LLM tutors accelerate convergence time while not crucially improving the RL model's best performance, as indicated in Table \ref{table:main_result} and Fig. \ref{fig:environments}. This outcome aligns with previous studies that LLM tutors accelerate convergence time \cite{ma2024explorllmguidingexplorationreinforcement, zhou2024largelanguagemodelpolicy}. Specifically, we find that the improvement in the RL models' performance notably varies in different configurations, indicating a highly sensitive optimization problem \cite{castillo2008sensitivity}. Generally, configurations with more complex environments usually show more noteworthy improvement over the baseline. This phenomenon is further emphasized when using the repeated advice mechanism. Moreover, DeepSeek outperforms the other LLM tutors, which can be attributed to its reasoning capabilities, which distinguish it from the other two tested LLMs. Nonetheless, due to the sensitivity of the optimization problem, some cases, such as the DQN together with DeepSeek for the Blackjack environment (with advice reuse) obtained just slightly better results compared to the baseline (0.6073 and 0.6216, respectively) and the DQN with LLaMA for the Connect Four environment (with advice reuse) obtained slightly worse results compared to the baseline (0.6824 and 0.7066, respectively). These outcomes emphasize that one cannot blindly use LLM-tutoring for RL problems and expect improvement in the RL's convergence, while such is obtained, on average.

On top of that, there is a weak correlation (\(r = 0.24, p > 0.05\)) between the size of an LLM and its effectiveness as a tutor to the RL models. Larger LLM models tend to yield better results, as seen with DeepSeek-R1 and Vicuna, while smaller models like LLaMA often underperform compared to the baseline, as revealed by \ref{fig:blackjack_dqn}. This suggests that larger LLMs, while still small compared to current state-of-the-art LLMs, may be more suitable for tutor roles in these settings, highlighting the importance of model size in such applications and aligning with the literature connecting models' expressiveness and performance \cite{exp_rl_llm_1,exp_rl_llm_2,exp_rl_llm_3}. Nonetheless, since this result turned out to be not statistically significant, it should be taken with a grain of salt. 

Notably from Fig. \ref{fig:environments} and Table \ref{table:main_result}, DQN outperforms the other two RL models for most configurations, in both the LLM-tutor and environments. A potential reason for the strong performance of LLM-assisted DQN agents is their ability to reduce the need for initial exploration. Traditional DQN relies on an exploration/exploitation strategy (e.g., \(\epsilon\)-greedy) to avoid local optima by taking random actions. However, with guidance from an LLM, the agent can bypass this random exploration phase, leveraging the model's knowledge to make more informed decisions early on. As a result, learning curves tend to show faster convergence and avoid the initial dip in performance typically seen in purely exploratory approaches (e.g. \cite{dqnHumanLevel}).

Despite the advantages of the LLM-tutoring in terms of the RL-steps, the time required to train an RL agent with an LLM tutor is considerably longer than when training without it, as indicated by Table \ref{table:timeSave}. This aligns with previous studies, \cite{cao2024sparserewardsenhancingreinforcement} utilized a cloud-based LLM and communicated with it through an HTTP API, which introduced rate limitations that slowed down the RL model's training. In our study, the time required for the LLM model to generate all of the tokens of its response was the main bottleneck holding the training speed back. Nevertheless, Table \ref{table:timeSave} exhibits that for relatively simple environments, the reduction in time for the RL model's training is considerable, especially when accounting for the extra time required when training with an LLM tutor. This effect was especially pronounced in experiments with DeepSeek-R1, as it includes its reasoning process in the response, significantly increasing the number of generated tokens and, consequently, the average response time.

Based on these findings, we offer the following practical considerations for integrating LLM tutors into reinforcement learning (RL) training workflows. First, given the observed sensitivity of performance gains to specific configurations, practitioners should carefully evaluate the suitability of LLM tutoring on a case-by-case basis, particularly in complex environments where the benefits appear more pronounced. Second, we recommend prioritizing larger LLMs that have undergone domain-specific fine-tuning \cite{Jeong_2024}, as these models consistently provide stronger guidance and accelerate convergence more reliably. However, it is crucial to weigh these benefits against the significantly increased inference time and computational overhead, especially with models that produce verbose outputs. To mitigate potential training delays, one can optimize LLM prompts for conciseness and explore solutions like faster local inference or knowledge distillation techniques.

This study is not without limitations. First, due to computational constraints, all results are derived from a small number of \textit{in silico} repetitions (\(N=3\)), which may introduce bias stemming from the inherent stochasticity of both RL algorithms and LLM outputs. Second, the quality and consistency of LLM-generated advice remain a challenge. The models were used in a zero-shot setting without fine-tuning, which likely limited their effectiveness. Although structured prompting was applied to encourage more reliable outputs, the non-deterministic nature of LLMs led to occasional invalid or syntactically incorrect suggestions, requiring prompt revisions and increasing training time. Third, the quality of LLM advice was not manually or systematically evaluated by human experts. Without human verification, it is challenging to gauge whether performance gains were due to genuinely helpful advice in the RL's policy level incidental correlations. For future work, we propose extending this investigation using more powerful, cloud-hosted LLMs, which may generate higher-quality and context-aware advice and thus better demonstrate the full potential of LLM tutoring with advice reusing, especially in more complex or higher-dimensional environments. Furthermore, exploring advanced advice reuse strategies beyond the basic reuse mechanism employed here, such as Q-change or decaying probability approaches \cite{10.5555/3398761.3398953}, could enhance training efficiency. Developing and evaluating novel reuse mechanisms tailored to specific advice types or environment dynamics also presents a promising venue for future research.

Taken jointly, this study demonstrates that LLMs have the potential to be effective tutors in a student-teacher relationship with RL models across various environments, increasing convergence time without affecting overall optimal performance. In particular, we find that an advice reuse mechanism in this context is able to marginally accelerate the observed effect while revealing context-dependent advantages. More refined exploration of this relationship, taking into account the RL, LLM, and environment properties, should further reveal the optimal design of this architecture.

\printbibliography

\end{document}